\documentclass{article}

\PassOptionsToPackage{numbers, compress}{natbib}

\usepackage[final]{neurips_2019_ml4ad}

\usepackage[utf8]{inputenc} 
\usepackage[T1]{fontenc}    
\usepackage{hyperref}       
\usepackage{url}            
\usepackage{booktabs}       
\usepackage{amsfonts}       
\usepackage{nicefrac}       
\usepackage{microtype}      

\usepackage{mathtools}        
\usepackage{longtable}      
\usepackage{listings}
\usepackage{xcolor}
\usepackage{amssymb}
\setcitestyle{numbers, open={[}, close={]}}

\DeclareMathOperator*{\argmax}{argmax}

\title{Multi-Agent Connected Autonomous Driving using Deep Reinforcement Learning}

\author{%
  Praveen Palanisamy\thanks{Microsoft AI + R \newline Code available at: \url{https://github.com/praveen-palanisamy/macad-gym}} \\
  \texttt{praveen.palanisamy@\{microsoft, outlook\}.com} \\
}

\begin{document}

\maketitle

\begin{abstract}
  The capability to learn and adapt to changes in the driving environment is crucial for developing autonomous driving systems that are scalable beyond geo-fenced operational design domains.
  Deep Reinforcement Learning (RL) provides a promising and scalable framework for developing adaptive learning based solutions.
  Deep RL methods usually model the problem as a (Partially Observable) Markov Decision Process in which an agent acts in a stationary environment to learn an optimal behavior policy.
  However, driving involves complex interaction between multiple,  intelligent (artificial or human) agents in a highly non-stationary environment.
  In this paper, we propose the use of Partially Observable Markov Games(POSG) for formulating the connected autonomous driving problems with realistic assumptions.
  We provide a taxonomy of multi-agent learning environments based on the nature of tasks, nature of agents and the nature of the environment to help in categorizing various autonomous driving problems that can be addressed under the proposed formulation.
  As our main contributions, we provide MACAD-Gym, a Multi-Agent Connected,
  Autonomous Driving agent learning platform for furthering research in this direction. Our MACAD-Gym platform provides an extensible set of Connected Autonomous Driving 
 (CAD) simulation environments that enable the research and development of Deep RL- based integrated sensing, perception, planning and control algorithms for CAD
 systems with unlimited operational design domain under realistic, multi-agent
 settings.
 We also share the MACAD-Agents that were trained successfully using the MACAD-Gym platform to learn control policies
for multiple vehicle agents in a partially observable, stop-sign controlled, 3-way urban intersection environment with raw (camera) sensor observations.

\end{abstract}

\section{Introduction}

Driving involves complex interactions between other
agents that is near-impossible to be exhaustively described through
code or rules. Autonomous driving systems for that reason cannot be pre-programmed with exhaustive rules to
cover all possible interaction mechanisms and scenarios
on the road. Learning agents can potentially discover
such complex interactions automatically through exploration and
evolve their behaviors and actions to be more
successful in driving based on their experiences gathered through 
interactions with the driving environment (over time and/or
in simulation). The Deep RL framework \cite{dqn} \cite{drl4ad} provides a scalable framework for developing adaptive, learning-based solutions for such problems.
But, it is hard to apply RL algorithms to live systems \cite{rlchallenges}, especially robots and safety-critical systems like autonomous cars and RL-based learning is not very sample efficient \cite{rlchallenges}
One way to overcome such limitations is by using realistic simulation environments to
train these agents and transfer the learned policy to the actual car. High-fidelity Autonomous driving
simulators like CARLA \cite{carla} and AirSim \cite{airsim} provide a simulation platform for
training Deep RL agents in singe-agent driving scenarios.

In single-agent learning frameworks, the interaction between other agents in the environment or
even the existence of other agents in the environment
is often ignored. In Multi-Agent learning frameworks, the
interaction between other agents can be explicitly modeled.

Connectivity among vehicles
are becoming ubiquitous and viable through decades of
research in DSRC and other vehicular communication
methods. With the increasing deployment of 5G infrastructure for 
connectivity and the increasing penetration
of autonomous vehicles with connectivity and higher levels of 
autonomy \cite{J3016_201806}, the need
for the development of methods and solutions that can utilize 
connectivity to enable safe, efficient, scalable and economically 
viable Autonomous Driving beyond Geo-fenced
areas has become very important to our transportation
system.

Autonomous Driving problems involve autonomous vehicles navigating safely and socially 
from their start location to the desired goal location in complex environments which 
involve multiple, intelligent actors whose intentions are not known by other actors.
Connected Autonomous Driving makes use of connectivity between vehicles (V2V), between
vehicles and infrastructure (V2I), between vehicles and pedestrians (V2P) and between
other road-users.

\begin{figure}
  \centering
  \includegraphics[width=\linewidth]{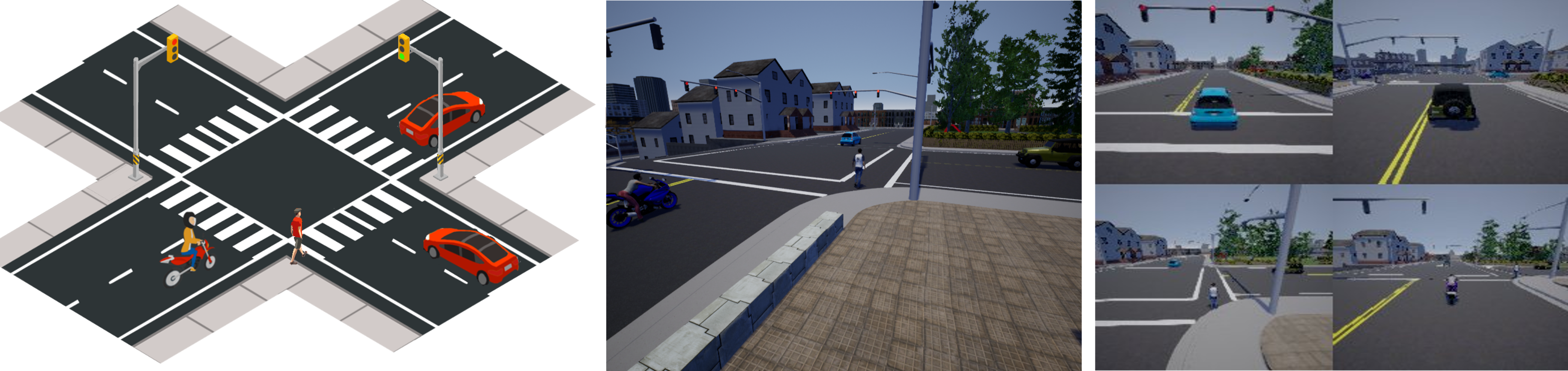}
  \caption{Figure shows a heterogeneous multi-agent learning environment created using MACAD-Gym. An overview of the scenario is shown in the left image. The middle image shows the simulated scenario and the right image shows tethered views of each agent's observation}
  \label{fig:macad-gym-intro-fig}
\end{figure}

CAD problems can be approached using homogeneous, communicating
multi-agent driving simulation environments for research and development of learning based
solutions. In particular, such a learning environment enables training and testing of RL
algorithms. To that end, in this paper,...
1. We propose the use of Partially Observable Markov Games for formulating the connected autonomous driving problems with realistic assumptions. 
2. We provide a taxonomy of multi-agent learning environments based on the nature of tasks, nature of agents and the nature of the environment to help in categorizing various autonomous driving problems that can be addressed under the proposed formulation.
3. We provide MACAD-Gym, a multi-agent learning platform with an extensible set of Connected Autonomous Driving (CAD) simulation environments that enable the research and development of Deep RL based integrated sensing, perception, planning and control algorithms for CAD systems with unlimited operational design domain under realistic, multi-agent settings.
4. We also provide MACAD-Agents, a set of baseline/starter agents to enable the
community to conduct learning experiments and train agents using the platform. The results of multi-agent policy learning by one of the provided baseline approach, trained in a partially observable, stop-sign controlled, 3-way urban intersection environment with raw, camera observations are summarized in \ref{sec:xp-n-conc}.
experimental results in a multi-agent settings with raw, simulated camera/sensor
observations to learn heterogeneous control
 policies to pass through a signalized, 4-way, urban intersection in a
 partially observable multi-agent, CAD environment with
 two cars, a pedestrian and a motor cyclist where all the actors are controlled by
 our MACAD-Agents.
Figure \ref{fig:macad-gym-intro-fig} depicts an overview of one of the MACAD environments released as a part of the MACAD-Gym platform.

The rest of the paper is organized as follows: We discuss how partially-observable
markov games (POMG) can be used to model connected autononomous driving problems
in \ref{sec:cadaspomg}. We then provide an intuitive classification of the tasks
and problems
in the CAD space in section \ref{sec:MACAD} and discuss the nomenclature of the MACAD-Gym environments
in section \ref{environment-naming-conventions}. We provide a quick overview of multi-agent deep RL algorithms in the context of CAD in section \ref{madrl4cad} and conclude with a brief discussion about the result obtained using MACAD-Agents 
in a complex multi-agent driving environment.

\section{Connected Autonomous Driving as Partially Observable Markov Games}
\label{sec:cadaspomg}

In single-agent learning settings, the interaction between the main
agent and the environment is modeled as part of a 
Markov Decision Process (MDP). Other agents (if) present are 
and treated to be part of the
environment irrespective of their nature (cooperative,
competitive), type (same/different as the main agent)
and sources of interactions with the main agent. Failing to account for the
presence of other intelligent/adaptive agents in the environment 
leads to conditions that violate the stationary
and Markov assumptions of the underlying learning
framework. In particular, when other intelligent agents
that are capable of learning and adapting their policies
are present, the environment becomes non-stationary.

\subsection{Formulation}
One way to generalize the MDP to account for multiple agents in multiple state configurations is using
Markov Games \cite{Littman:1994} which re defines the game-
theoretic stochastic games \cite{shapley1953stochastic} formulation in the 
reinforcement learning context.
In several real-world multi-agent problem domains like autonomous driving, assuming that each agent can observe
the complete state of the environment without uncertainty is unrealistic, partly due to the nature of the sensing capabilities present in the vehicle (actor), the physical embodiment of the agent.
Partially Observable Stochastic Games (POSG) \cite{Emery-Montemerlo:2004} extend stochastic games to problems with partial observability.
In the same vein as Markov Games, we re-define POSG in the context of
reinforcement learning as Partially Observable Markov Games,
POMG in short, as a tuple $\langle \mathcal{I}, \mathcal{S},
\mathcal{A},\mathcal{O}, \mathcal{T}, \mathcal{R} \rangle$ in which,

$\mathbf{I}$ is a finite set of actors/agents

$\mathcal{S}$ is the finite set of states

$\mathcal{A} = \times_{i \in I}\mathcal{A}_i $ is the set of joint actions where
$\mathcal{A}_i$ is the set of actions available to agent \(i\).

$\mathcal{O} = \times_{i \in I} \mathcal{O}_i$  is the set of join observations where
$\mathcal{O}_i$ is the set of observations for agent i.

$\mathcal{T} = \mathcal{P}(\mathbf{s'}, \mathbf{o} | \mathbf{s}, \mathbf{a})$ is the Markovian state transition and observation
probability that taking a joint action $\mathbf{a}=\langle a_1, ... a_n \rangle$ in state $s$ results in a transition to state $s'$ with a join observation $\mathbf{o}= \langle o_1, ... o_n \rangle$

$\mathcal{R}_i:
\mathcal{S} \times \mathcal{A} \to \mathbb{R}$ is the reward
function function for agent \(i\)

At each time step \(t\), the environment emits a joint observation
\(\mathbf{o} = \langle o_1, \cdots, o_n \rangle\) from which each agent
\(i\) directly observes its component \(o_i \in \mathcal{O}_i\) and takes action
\(a_i \in A_i\) based on some policy
\(\pi_i : \mathcal{O}_i \times \mathcal{A}_i \to [0, 1]\) and receives a
reward \(r_i\) based on the reward function
$\mathcal{R}_i$ . 

Note that, the above formulation is equivalent to a POMDP when \(n=1\)
(single-agent formulation).

While a POSG formulation of the autonomous driving problem enables one
to approach the problem with out making unrealistic assumptions, it does
not enable computationally tractable methodologies to solve the
problem except under simplified, special structures and
assumptions like two-player zero-sum POSGs. In the next section, we
 discuss the practical usage in the CAD domain. 

\subsection{Practical usage in Connected Autonomous Driving}
\label{pomg-usage-in-ad}
The availability of a communication (whether through explicit communication semantics or implicitly through augmented actions)
channel between the agents (and/or the env) in the CAD domain, enable
the sharing/transaction of local information (or private beliefs) that can provide
information about some (or whole) subset of the state
which is locally observable by other agents, make solutions
computationally tractable even as the size of the
problem (eg: num agents) increases. We realize that such
a transaction of local information would give rise to issue
of integrity, trust and other factors.
The interaction between different agents and the nature
of their interaction can be explicitly modeled using a
communication channel. In the absence of an explicit
communication channel, the incentives for the agents to
learn to cooperate or compete, depends on their reward
functions. The particular case in which all the agents
acting in a partially observable environment share the
same reward function, can be studied under the DEC-
POMDP \cite{oliehoek2012decentralized} formulation. But not all problems in
CAD have the agent’s rewards completely aligned (or
completely opposite).

We consider Multi-Agent Driving environments with $n$ 
actors, each controlled by an agent, indexed by $i$.
At any given time, $t$, the state of the agent $i$ is defined to be 
the state of the actor under its control and it is 
represented as ${s}_{i} \in \mathcal{S}$ where $\mathcal{S}$ is the 
state space.
The agent can choose an action $a_i \in \mathcal{A}_i$ , where 
$\mathcal{A}_i$ is the action space of agent $i$ which, could be different for different actors.
While the environment is non-Markov from each of the agent's point of view, the driving world as a whole is assumed to be Markov i.e, given 
the configuration of all the $n$ actors at time $t$ : $\mathbf{s}=[s_1, ... s_n]$ 
, their actions $\mathbf{a}= [a_1,... a_n]$, 
and the state of the environment $E$, 
the evolution of the system is completely determined by the 
conditional transition probability $\mathcal{T}(\mathbf{s'}, \mathbf{o}, E' | \mathbf{s}, E, \mathbf{a})$.
This assumption allows us to apply and scale distributed RL algorithms that are developed for the single-agent
MDPs to the Multi-Agent setting.

The explicit separation of the join state of the agents $\mathbf{s}_i$ from the state of the environment $E$ at time $t$ in the driving world, facilitates agent implementations to learn explicit models for the environment, in addition to learning models for other agents or the world model as a whole.

Under the proposed formulation for multi-agent CAD, at 
every time step $t$, each actor (and hence the agent) receives an 
observation $o_i$, based on its state $s_i$ and the (partial) state of the environment $E_i$ and possibly,
(partial) information $I(\mathbf{s_{-i}}, E_{-i})$ about the state of other 
agents $\mathbf{s_{-i}}= [s_j]_{j\ne i}$ and the state of the environment $E_{-i}$ that is not directly observable. 

The observation $o_i$ can be seen as some degraded function $\phi(s_i, E)$ of the full 
state of agent i . In reality, the nature and the degree of degradation arises 
from the sensing modalities 
and the type of sensors (camera, RADAR, LIDAR, GPS etc.) available 
to the vehicle actors. 
Connectivity through IEEE 802.11 based Dedicated Short-Range Communications (DSRC) \cite{dsrc} or
cellular modems based C-V2X \cite{cv2x} enables the availability of 
the information $I(\mathbf{s_{-n}}, E_{-i})$ about other agents and non-observable parts of the environment through Vehicle-to-Vehicle (V2V), Vehicle-to-Infrastructure (V2I) or Vehicle-to-Anything (V2X) communication.


The goal of each agent is to take actions $a_i$ for the vehicle 
actor that is under its control based on its local state $[o_i, I(\mathbf{s_{-i}, E_{-i}})]$
in order to maximize its long term cumulative reward over a time horizon T with a 
discount factor of $\gamma$.

\section{Multi-Agent Connected Autonomous Driving Platform}
\label{sec:MACAD}
The connected-autonomous driving domain poses several problems which can be 
categorized into sensing, perception, planning or control. Learning algorithms can be used to solve the tasks in an integrated/end-to-end fashion \cite{e2e} \cite{e2eaux} or in an isolated approach for each driving task like intersection-driving \cite{occluded-intx} \cite{intx} and lane-changing \cite{lc1}.

Driving tasks falling under each of the above categories can be further divided and approached, depending on the combination of the nature of the tasks, the nature of the environments and the nature of the agents. The following subsections provide a brief discussion
on such a classification of multi-agent environments that are supported on the MACAD-Gym platform, to enable the development of solutions for various tasks in the 
the CAD domain.
\subsection{Nature of tasks}
The nature of the task in a driving environment is determined based on the desired
direction of focus of the task specified through the design of experiments. 
\paragraph{Independent}
Multi-agent driving environments in which each actor is self-interested/selfish and has
its own, often unique objective, fall under this category. One way to model such setup
is by treating the environment to be similar to a single-agent environment with all the
actors apart from the host actor are treated to be be part of the environment. Such
environments help in developing non-communicating agents that doesn't rely on explicit
communication channels. Such agents will benefit from agents modeling agents \cite{modelotheragents}.
\paragraph{Cooperative}
Cooperative CAD environments help in developing agent algorithms that can learn near-globally optimal policies for all the driving agents that act as a cooperative unit. Such environments help in developing agents that learn to communicate \cite{l2comm} and benefit from learning to cooperate \cite{coop}. This type of environments will enable development of efficient fleet of vehicles that cooperate and communicate with each other to reduce congestion, eliminate collisions and optimized traffic flows.
\paragraph{Competitive}
Competitive driving environments allow the development of agents that can handle extreme driving scenarios like road-rages. The special case of adversarial driving can be 
formulated as a zero-sum stochastic game, which can be cast as a MDP and solved which has useful properties and properties and results including: value Iteration, unique solution to Q*, independent computation of policies and representation of policies using Q functions as discussed in \cite{zero-sum}.
Agents developed in competitive environments can be used for law enforcement and or other use cases including the development of strong adversarial driving actors to help improve handling capabilities of driving agents.
    
\paragraph{Mixed}
\label{nature-of-env-mixed}
Some tasks that are designed to be of a particular nature may still end up facilitating approaches that stretch the interaction to other types of tasks. For example, an agent operating in an environment on a task which is naturally (by
design) an independent task can learn to use mixed strategies of being cooperative at
times and being competitive at times in order to maximize it's own rewards. Emergence
of such mixed strategies \cite{mixed} is another interesting research area supported in MACAD-Gym, that can
lead to new traffic flow behaviors.

\subsection{Nature of agents/actors}
\paragraph{Homogeneous}
When all the road actors in the environment belong to one class of actors (eg. only cars or only motor-cyclists), the action space of each actor can be the same and the interactions are limited to be between a set of homogeneous driving agents.
\paragraph{Heterogeneous}
Depending on the level of detail in the environment representation (a traffic light could be represented as an intelligent actor), majority of autonomous driving tasks involve interaction between a heterogeneous set of road actors. 
\paragraph{Communicating}
Actors that are capable of communicating (through direct or indirect channels \cite{Panait:2005} with
other actors through Vehicle-to-Vehicle (V2V) communication channels can help to
increase information availability in partially-observable environments. Such
communication capabilities allow for training agents with data augmentation wherein the
communication acts as a virtual/shared/crowd-sourced sensor.
Note that, Pedestrian (human) agents can be modeled as communicating agents that use (hand and body) gestures transmit information and can receive information using visual (external display/signals on cars, Traffic signals etc) and auditory (horns, etc). 

\paragraph{Non-communicating}
 While the environment provides or allows for a communication channel, if an agent is not capable of communicating/making-use-of-the-communication channel by virtue of the nature of the actor, it is grouped under this category. Example include vehicle actors that have no V2X communication capability. 
  
\subsection{Nature of environments}
\paragraph{Full/partial observability} In order for an environment to be fully
observable, every agent in the environment should be able to observe the complete state
of the environment at every point in time. Driving environments under realistic assumptions are partially-observable environments. The presence of connectivity (V2V, V2X/cloud) in CAD environments make the problems in PO environments more tractable.
\paragraph{Synchronous/Asynchronous} In a synchronous environment, all
the actors are required to take an action in a time synchronous manner. Whereas, in
asynchronous environments, different actors can act at different frequencies.
\paragraph{Adversarial}If there exists any environmental factor/condition that can stochastically impair the ability of the agents in the environment to perform at their full potential, such cases are grouped under adversarial environments. For example, the V2X communication medium can be perturbed/altered by the environment,which enables the study of the robustness of agents under adversarial attacks. Bad weather including snowy, rainy or icy conditions also can be modeled and studied under adversarial environments. Injection of "impulse"/noise that are adversarial in nature help in validating the reliability of agent algorithms.

Table \ref{table:supported-MA-env} in appendix \ref{appendix:B} shows a short list of environments that are supported by the MACAD-Gym platform. 
\subsection{MACAD-Gym Environment Naming Conventions}\label{environment-naming-conventions}

A naming convention that conveys the environment type, nature of the
agent, nature of the task, nature of the environment with version
information is proposed. The naming convention follows the widely used
convention in the community introduced in \cite{gym} but, has been
extended to be more comprehensive and to accommodate more complex
environment types suitable for the autonomous driving domain.

The naming convention is illustrated below with
\textbf{HeteCommCoopPOUrbanMgoalMAUSID} as the example:

\newcommand{\upenum}[2]{\overset{\overset{#2}{\uparrow}}{#1}}
\newcommand{\downenum}[2]{\underset{\underset{#2}{\downarrow}}{#1}}

\[\downenum{\underbrace{
\upenum{\overbrace{\text{Hete}}}{\substack{\{\text{Hete}, \\ \text{Homo}\}}}
\upenum{\overbrace{\text{Comm}}}{\substack{\{\text{Comm}, \\ \text{Ncom\}}}}
}}{\text{\normalsize Nature of Agents}}
\downenum{\underbrace{\upenum{\overbrace{\text{Coop}}}{\substack{\{Inde, \\ Coop, \\ Comp, \\Mixd\}}}}}{{\text{\normalsize Nature of tasks}}}
\downenum{\underbrace{\upenum{\overbrace{\text{PO}}}{\substack{\{PO, \\ FO\}}}
\upenum{\overbrace{\text{Urban}}}{\substack{\{Bridg, \\ Freew, \\Hiway, \\ Intrx, \\Intst, \\ Rural, \\ Tunnl, \\Urban\}}}
\upenum{\overbrace{\text{Mgoal}}}{\substack{\{\null, \\ Advrs, \\ Async, \\ Mgoal, \\Synch, \\... \}}}
\upenum{\overbrace{\text{MA}}}{\substack{\{MA, \\ SA\}}}
}}{{\text{\normalsize Nature of environment}}}
\upenum{\overbrace{\text{USID}}}{\substack{Unique \\ Scenario \\ ID}}
\text{-}\downenum{\underbrace{\text{v0}}}{\text{\normalsize version}}\]

A few example environments that are part of the initial Carla-Gym
platform release are listed in Appendix \ref{appendix:B}.

The above description summarizes the naming convention to accommodate various types of driving environments with an
understanding that several scenarios and their variations can be created
by varying the traffic conditions, speed limits and behaviors of other
(human-driven, non-learning, etc) actors/objects in each of the environments. The
way these variations are accommodated in the platform is by using an
Unique Scenario ID (USID) for each variation in the scenario. 
The "version" string allows versioning each scenario variation when changes are made to the reward function and/or observation and actions spaces.

\section{Multi-Agent Deep Reinforcement Learning For Connected Autonomous Driving}
\label{madrl4cad}

In the formulation presented in section \ref{sec:cadaspomg}, formally, the goal of each agent is to maximize  the expected value of its long-term future reward given by the following objective function:

\begin{equation} \label{eq:local_obj}
    J_i(\pi_i, \boldsymbol{\pi_{-i}}) = \mathbb{E}_{\pi_i ,\boldsymbol{\pi_{-i}}}[R_i] 
     = \mathbb{E}_{\pi_i, \boldsymbol{\pi_{-1}}} [\sum^T_{t=0}\gamma^t r_i(\boldsymbol{s}, a_i)] 
\end{equation}

Where $\boldsymbol{\pi}_{-i}= \prod_j{\pi_j(\mathbf{s}, a_j), {j\ne i}}$ is the set of policies of agents other than agent $i$.
In contrast to the single-agent setting, the objective function of an agent in the multi-agent setting depends on the policies of the other agents. 

\subsection{Value Based Multi-Agent Deep Reinforcement Learning}

\begin{equation}
    V_i^{\boldsymbol{\pi}}(s) =  \sum_{\mathbf{a} \in \mathbf{A}} \boldsymbol{\pi}(s, \mathbf{a}) \sum_{s' \in \mathbf{S}}T(s, a_i, \mathbf{a_{-i}}, s')[R(s, a_i, \mathbf{a_{-i}}, s') + \gamma V_i^{\boldsymbol{\pi}}(s')]
\end{equation}

where, $s= (\mathbf{S}^t, E^t)$,
$\mathbf{a} = (a_i, \mathbf{a_{-i}})$, 
$\boldsymbol{\pi}(s, \mathbf{a}) = \prod_j{\pi_j(s, a_j)}$

The optimal policy is a best response dependent on the other agent's policies,

\begin{equation}
\pi_i^*(s, a_i, \boldsymbol{\pi}_{-i}) = {\argmax_{\pi_i}} V_i^{(\pi_i, \boldsymbol{\pi}_{-i})}(s)
\end{equation}

\begin{equation}
    = \argmax_{\pi_i}\sum_{\mathbf{a} \in A}\pi_i(\mathbf{s}, a_i) \boldsymbol{\pi}_{-i}(\mathbf{s}, \mathbf{a}_{-i})\sum_{s' \in \mathcal{S}}\mathcal{T}(\mathbf{s}, a_i, \mathbf{a}_{-i}, s') 
    [R(\mathbf{s}, a_i, \mathbf{a}_{-i}, s') + \gamma V_i^{(\pi_i, \boldsymbol{\pi}_{-i})}(s')]
\end{equation}

Computing the optimal policy under this method requires $\mathcal{T}$, the transition model of the environment to be known. The state-action value function $Q_i^*(s_i, a_i| \boldsymbol{\pi}_{-i})$ is presented in appendix \ref{appendix:B}. 

\subsection{Policy Gradients} 
If $\boldsymbol{\theta} = \{\theta_1, \theta_2, ..., \theta_N\}$ represents the parameters of the policy $\boldsymbol{\pi} = \{\pi_i, \pi_2, ...,\pi_N\}$, The gradient of the objective function (equation \ref{eq:local_obj}) w.r.t the policy parameters can be written as:
\begin{equation}
    \nabla_{\theta_i}J_i(\pi_i, \boldsymbol{\pi}_{-i}) = \mathbb{E}_{\boldsymbol{S}^t, E^t\sim {p^{\boldsymbol{\pi}}}}
\end{equation}

\subsection{Decoupled Actor - Learner architectures}

For a given CAD situation with N homogeneous driving agents, the globally optimal solution is the policy that maximizes the  following objective:
\begin{equation}
\label{eq:global_obj}
    \mathbb{E}_\pi[\sum_{i=1}^N R_i] =\sum_{i=1}^N\mathbb{E}_{\pi_i, \boldsymbol{\pi_{-i}}}[R_i]
\end{equation}

The straight-forward approach to optimize for the global objective (equation \ref{eq:global_obj}) amounts to finding the globally optimal policy:
\begin{equation}
    \boldsymbol{\pi}^{g^*} = \argmax_{\pi} \sum_{i=1}^N J_i ( \pi_i, \boldsymbol{\pi_{-i}})
\end{equation}
However, this approach requires access to policies of all the agents in the environment.

\begin{figure}
  \centering
  \includegraphics[width=\linewidth]{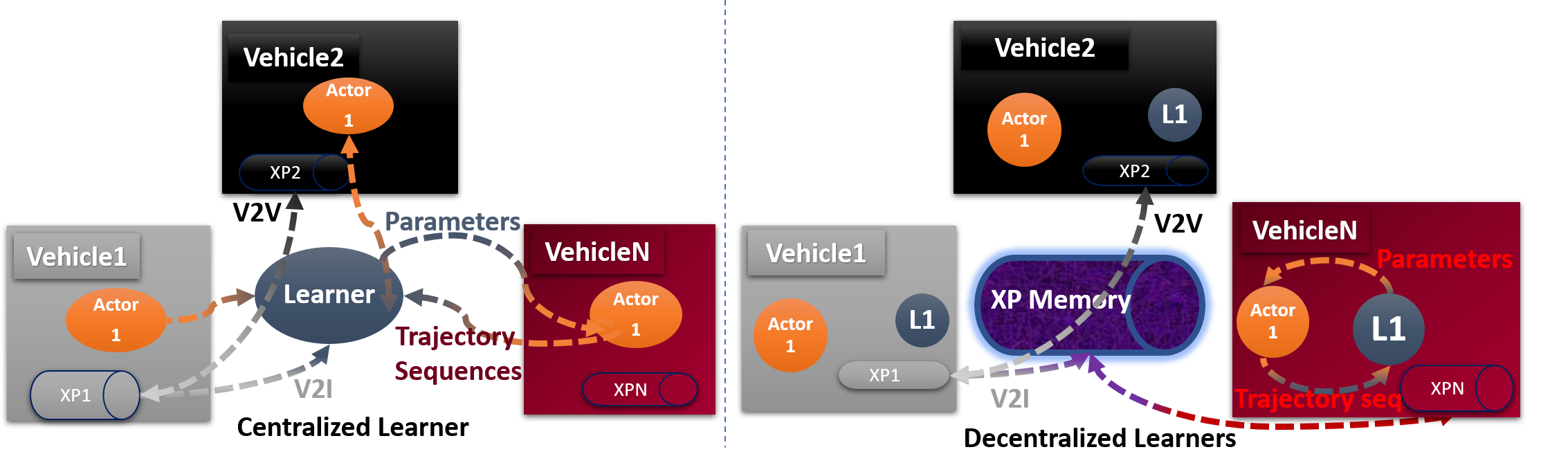}
  \caption{Centralized learner (left) and decentralized learner(right) architecture for connected-autonomous driving}
  \label{fig:centralized-decentralized}
\end{figure}

\paragraph{Centralized Learners}
Figure \ref{fig:centralized-decentralized} (left) depicts decoupled actor-learner
architecture with a centralized learner which can be used to learn a
globally-optimal driving policy $\boldsymbol{\pi}^{g^*}$.

\paragraph{Decentralized Learners} In the most general case of CAD, each driving
agent follows it's own policy that is independent of the other agent's driving
policy. Note that this case can be extended to cover those situations where  some
proportion of the vehicles are driven by humans who have their own intentions and policies.

Each agent can independently learn, to find policies that optimize their local objective function (equation \ref{eq:local_obj}). One such architecture for CAD is shown in figure \ref{fig:centralized-decentralized} (right).

\paragraph{Shared Parameters} With connectivity as in CAD, some are all the parameters of each agent's policy can be shared with one another. Such parameter sharing between driving agents can be implemented with both centralized and decentralized learner architectures.

\paragraph{Shared Observations} Sharing observations from the environment with other agents via communication medium, reduces the gap between the observation $o_i$ and the true state $\langle s_i, E \rangle$ and can drive the degradation function $\phi(s_i, E)$ (discussed in section \ref{pomg-usage-in-ad}) to Identity (no degradation).

\paragraph{Shared Experiences}This enables collective experience replay which can theoretically lead to gains in a way similar to distributed experience replay \cite{apex} in single-agent setting.

\paragraph{Shared policy} If all the vehicles follow the same policy $\pi^l$, it follows from equation \ref{eq:local_obj} that the learning objective for each of the agents can be simplified, resulting in an identical and equal definition:
\begin{equation}
    J(\pi^l) = \mathbb{E}_{\pi^l}[\sum_{t=0}^T \gamma^t r_i(\mathbf{S}^t, E^t, a_i^t)]
\end{equation}
In this setting, challenges due to the non-stationarity of the environment is subsided due to the perfect knowledge about other agent's policies. 
In practice this case is of use in autonomous fleet operations in controlled environments where all the autonomous driving agents can be designed to follow the same policy

\section{Experiments and Conclusion}
\label{sec:xp-n-conc}
\begin{figure}
  \centering
  \includegraphics[width=\linewidth]{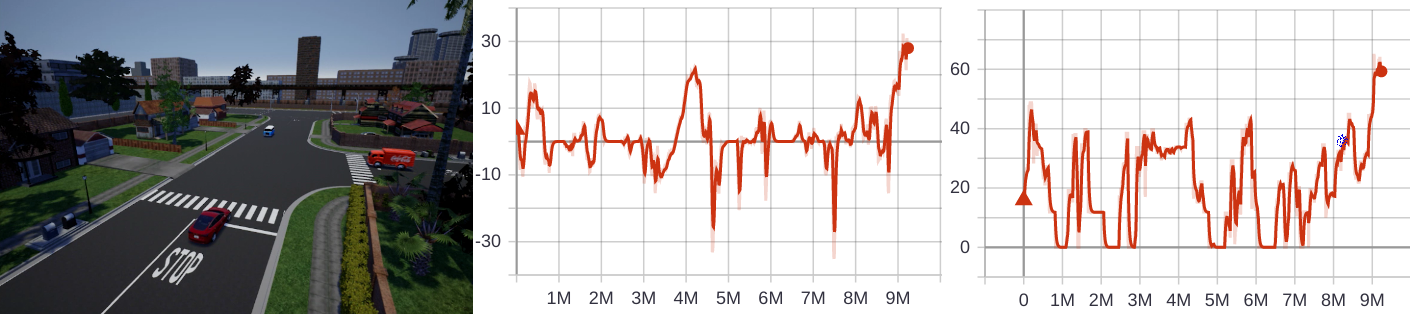}
  \caption{Figure shows a start state in the `HomoNcomIndePOIntrxMASS3CTWN3-v0`
  environment (left) and the cumulative mean episode rewards (middle) and the
  cumulative max episode rewards (right) obtained by the 3-agent system.}
  \label{fig:experiment-setup-res}
\end{figure}

We trained MACAD-Agents in the HomoNcomIndePOIntrxMASS3CTWN3-v0 environment,
which is a stop sign-controlled urban intersection environment with homogeneous, 
non-communicating actors, each controlled using the IMPALA \cite{impala}
agent architecture. The actors car1(red cola van), car2(blue minivan), car3 (maroon sedan)
learn a reasonably good driving policy to completely cross the intersection without 
colliding and within the time-limit imposed by the environment.
The environment is depicted in Figure \ref{fig:experiment-setup-res} along with the
cumulative mean and max rewards obtained by the 3-agent system. Complete details about the
experiment including agent-wise episodic rewards are presented in appendix \ref{appendix:C}. 

To conclude, we described a POSG formulation and discussed how CAD problems can be studied under such a formulation for various categories of tasks. We presented the opensource MACAD-Gym platform and the starter MACAD-Agents to help researchers to explore the CAD domain using deep RL algorithms
We also provided preliminary experiment results that validated the MACAD-Gym platform by conducting a starter experiment with the
MACAD-Agents in a multi-agent driving environment and discussed the results showing the
ability of the agents to learn independent vehicle control policies from high-dimensional
raw sensory (camera) data in a partially-observed, multi-agent simulated driving
environment. 
The MACAD-Gym platform enables training driving agents for several challenging autonomous driving problems. As a future work, we will develop a benchmark with a standard set of environments that can serve as a test-bed for evaluating machine-learning-based CAD driving agent algorithms.


\clearpage
\appendix
\section{AppendixA}
\label{appendix:A}

\subsection{Actor - Agent disambiguation}
In the context of this paper, to avoid any ambiguity between the 
usage of the terms, we consider an \textit{actor} to be a physical 
entity with some form of embodied intelligence, acting in an environment.
We consider an \textit{agent} to be a (software/algorithm) entity
that provides the intelligence to an actor. The agent can 
learn and/or adapt based on its interaction with the environment by controlling the (re)action of the actor.

\subsection{Available and supported environments in the MACAD-Gym platform}
A short-list of CAD environments made available on the MACAD-Gym platform are listed below with a brief description:

\begin{longtable}[]{@{}cc@{}}
\caption{An example list of environments that are part of the carla-gym
platform with a description to explain the naming conventions to make it
easy for the community to add new classes of environments.
\label{table:env_name_desc}}\tabularnewline
\toprule
\begin{minipage}[b]{0.35\columnwidth}\centering\strut
Environment Name\strut
\end{minipage} & \begin{minipage}[b]{0.59\columnwidth}\centering\strut
Description\strut
\end{minipage}\tabularnewline
\midrule
\endfirsthead
\toprule
\begin{minipage}[b]{0.35\columnwidth}\centering\strut
Environment Name\strut
\end{minipage} & \begin{minipage}[b]{0.59\columnwidth}\centering\strut
Description\strut
\end{minipage}\tabularnewline
\midrule
\endhead

\begin{minipage}[t]{0.35\columnwidth}\centering\strut
HomoNcomIndePOIntrxMASS3CTwn3-v0\strut
\end{minipage} & \begin{minipage}[t]{0.59\columnwidth}\raggedleft\strut
\textbf{Homo}geneous \textbf{N}on\textbf{com}municating,
\textbf{Inde}pendent, \textbf{P}artially-\textbf{O}bservable  
\textbf{Intersection}, \textbf{M}ulti-\textbf{A}gent
\textbf{Env}ironment with a \textbf{S}top \textbf{S}ign, \textbf{3C}ar
 scenario in \textbf{Town03}
\textbf{v}ersion\textbf{-0} \strut
\end{minipage}\tabularnewline

\begin{minipage}[t]{0.35\columnwidth}\centering\strut
HeteCommIndePOIntrxMAEnv-v0\strut
\end{minipage} & \begin{minipage}[t]{0.59\columnwidth}\raggedleft\strut
\textbf{Hete}rogeneous \textbf{Comm}unicating, \textbf{Inde}pendent,
\textbf{P}artially\textbf{O}bservable\textbf{Intersection}
\textbf{M}ulti-\textbf{A}gent \textbf{Env}ironment \textbf{v}ersion
\textbf{0}\strut
\end{minipage}\tabularnewline

\begin{minipage}[t]{0.35\columnwidth}\centering\strut
HeteCommCoopPOUrbanMAEnv-v0\strut
\end{minipage} & \begin{minipage}[t]{0.59\columnwidth}\raggedleft\strut
\textbf{Hete}rogeneous \textbf{Comm}unicating, \textbf{Coop}erative,
, \textbf{P}artially-\textbf{O}bservable, \textbf{Urban}
\textbf{M}ulti-\textbf{A}gent \textbf{Env}ironment \textbf{v}ersion
\textbf{0}\strut
\end{minipage}\tabularnewline

\begin{minipage}[t]{0.35\columnwidth}\centering\strut
HomoNcomIndeFOHiwaySynchMAEnv-v0\strut
\end{minipage} & \begin{minipage}[t]{0.59\columnwidth}\raggedleft\strut
\textbf{Homo}geneous \textbf{N}on\textbf{com}municating,
\textbf{Inde}pendent, 
\textbf{F}ully-\textbf{O}bservable, \textbf{Hi}gh\textbf{way}, \textbf{M}ulti-\textbf{A}gent
\textbf{Env}ironment \textbf{v}ersion\textbf{-0}\strut
\end{minipage}\tabularnewline

\bottomrule
\end{longtable}

\label{table:supported-MA-env}
\begin{longtable}[]{@{}llrrr@{}}
\caption{Supported MA environment types{}} \tabularnewline
\toprule
& & \textbf{Independent} & \textbf{Cooperative} &
\textbf{Competitive}\tabularnewline
\midrule
\endfirsthead
\toprule
& & \textbf{Independent} & \textbf{Cooperative} &
\textbf{Competitive}\tabularnewline
\midrule
\endhead
\textbf{Homogeneous} & Communicating & \checkmark & \checkmark & \checkmark\tabularnewline
& Non-Communicating &  \checkmark & \checkmark & \checkmark\tabularnewline
\textbf{Heterogeneous} & Communicating & \checkmark& \checkmark& \checkmark\tabularnewline
& Non-Communicating &  \checkmark & \checkmark& \checkmark\tabularnewline
\bottomrule
\end{longtable}
\section{Appendix B}
\label{appendix:B}

\subsection{State-action value function}
In a fully-observable, Single-Agent setting, the optimal action-value function $Q^*(s, a)$ can be estimated using the following equation:
\begin{equation} \label{eq:q*}
    Q^*(s, a) = \mathbb{E}_{s'}[r + \gamma max_{a'}Q^*(s', a') | s, a].
\end{equation}

DQN \cite{dqn} uses a neural network to represent the action-value function parametrized by $\theta$, $Q(s, a; \theta)$.
The parameters are optimized iteratively by minimizing the Mean Squared Error (MSE) between the Q-network and the Q-learning target using Stochastic Gradient Descent with the loss function given by: 
\begin{equation} \label{eq:dqn_loss}
    \mathcal{L}(\theta) = \mathbb{E}_{s, a, r, s' \sim \mathcal{D}}[(r + \gamma max_a' Q(s', a';\theta^-) - Q(s, a; \theta))^2]
\end{equation}

where $\mathcal{D}$ is the experience replay memory containing $(s_t, a_t, r_t, s')$ tuples.

For a fully-observable, Multi-Agent setting, the optimal action-value function $Q^*(s, a | \pi_{-i})$ can be estimated using the following equation:
\begin{equation}
\begin{split}
    Q^*(s, a | \pi_{-i}) = \sum_{a_{-i}}\pi_{-i}(a_{-i}, s) \mathbb{E}_{s'}[r_i(s, a, s') \\
    + \gamma \mathbb{E}_{a'}[Q^*(s', a' | \pi_{-i})]]
\end{split}
\end{equation}

where $\pi_{-i}$ is the joint policy of all agents other than agent $i$, $s$ and $a$ are the state and action of agent $i$ at time-step $t$ and $s'$, $a'$ are the state and action of agent $i$ at time-step $t+1$.

Independent DQN \cite{coop} extends DQN to cooperative, fully-observable Multi-Agent setting, applied to a two-player pong environment, in which all agents independently learn and update their own Q-function $Q_i(s, a_i; \theta_i)$. 

Deep Recurrent Q-Network \cite{DRQN} extends DQN to the partially-observable Single-Agent setting by replacing the first post-convolutional fully-connected layer with a recurrent LSTM layer to learn Q-functions of the form: $Q(o_t, h_{t-1}, a; \theta_i)$ that generates $Q_t$ and $h_t$ at every time step, where $o_t$ is the observation and $h_{t}$ is the hidden state of the network. 

For a Multi-Agent setting, 
\begin{equation}
    Q_i^*(s_i, a_i| \boldsymbol{\pi}_{-i}) = \sum_{\mathbf{a}_{-i}} \boldsymbol{\pi}_{-i}(\mathbf{a}_{-i} | s) \mathbb{E}_{s_i}[ r_i(\mathbf{s}, E, \mathbf{a}) + \gamma \mathbb{E}_{a'}[Q_i(s_i, a_i | \boldsymbol{\pi}_{-i})]
\end{equation}
\section{Experiment description}
\label{appendix:C}

\lstset{
    string=[s]{"}{"},
    stringstyle=\color{blue},
    comment=[l]{:},
    commentstyle=\color{black},
}
\subsection{Environment description}
 The `HomoNcomIndePOIntrxMASS3CTWN3-v0` follows the naming convention discussed in \ref{environment-naming-conventions} and refers to a homogeneous, non-communicating, independent, partially-observable multi-agent, intersection environment with stop-sign controlled intersection scenario in Town3. The SUID is '' (empty string) and the version number is `v0`.
 The environment has 3 actors as defined in the scenario description (\ref{actor-goals-SSUI3C}). The description of each actor, their goal coordinates and their reward functions are described below:

\subsubsection{Actor description}

\begin{lstlisting}
{
  "actors": {
    "car1": {
      "type": "vehicle_4W",
      "enable_planner": true,
      "convert_images_to_video": false,
      "early_terminate_on_collision": true,
      "reward_function": "corl2017",
      "scenarios": "SSUI3C_TOWN3_CAR1",
      "manual_control": false,
      "auto_control": false,
      "camera_type": "rgb",
      "collision_sensor": "on",
      "lane_sensor": "on",
      "log_images": false,
      "log_measurements": false,
      "render": true,
      "x_res": 168,
      "y_res": 168,
      "use_depth_camera": false,
      "send_measurements": false
    },
    "car2": {
      "type": "vehicle_4W",
      "enable_planner": true,
      "convert_images_to_video": false,
      "early_terminate_on_collision": true,
      "reward_function": "corl2017",
      "scenarios": "SSUI3C_TOWN3_CAR2",
      "manual_control": false,
      "auto_control": false,
      "camera_type": "rgb",
      "collision_sensor": "on",
      "lane_sensor": "on",
      "log_images": false,
      "log_measurements": false,
      "render": true,
      "x_res": 168,
      "y_res": 168,
      "use_depth_camera": false,
      "send_measurements": false
    },
    "car3": {
      "type": "vehicle_4W",
      "enable_planner": true,
      "convert_images_to_video": false,
      "early_terminate_on_collision": true,
      "reward_function": "corl2017",
      "scenarios": "SSUI3C_TOWN3_CAR3",
      "manual_control": false,
      "auto_control": false,
      "camera_type": "rgb",
      "collision_sensor": "on",
      "lane_sensor": "on",
      "log_images": false,
      "log_measurements": false,
      "render": true,
      "x_res": 168,
      "y_res": 168,
      "use_depth_camera": false,
      "send_measurements": false
    }
  }
}
\end{lstlisting}

\subsubsection{Goals}
\label{actor-goals-SSUI3C}
The goal of actor car3 (maroon sedan) is to successfully cross the intersection by going straight. The goal of actor car1 (Red cola van) is to successfully cross the intersection by taking a left turn. The goal of actor car2 (blue minivan) is to successfully cross the intersection by going straight. For all the agents, successfully crossing the intersection amounts to avoiding collisions or any road infractions and reaching the goal state within the time-limit of one episode.

The start and goal coordinates of each of the actors in CARLA Town03 map is listed below for ground truths:

\begin{lstlisting}

SSUI3C_TOWN3 = {
    "map": "Town03",
    "actors": {
        "car1": {
            "start": [170.5, 80, 0.4],
            "end": [144, 59, 0]
        },
        "car2": {
            "start": [188, 59, 0.4],
            "end": [167, 75.7, 0.13],
        },
        "car3": {
            "start": [147.6, 62.6, 0.4],
            "end": [191.2, 62.7, 0],
        }
    },
    "weather_distribution": [0],
    "max_steps": 500
}
\end{lstlisting}

\subsection{Observation and Action spaces}
The observation for each agent is a 168x168x3 RGB image captured from the camera mounted on the respective actor that the agent is controlling.
The action space is Discrete(9). The mapping between the discrete actions and the vehicle control commands (steering, throttle and brake) are provided in table \ref{table:action-space-convertion}

\begin{table}
\label{table:action-space-convertion}
\centering
\begin{tabular}{lll}
\hline
Action & {[}Steer, Throttle, Brake{]} & Description \\
\hline
0 & {[}0.0, 1.0, 0.0{]} & Accelerate) \\
1 & {[}0.0, 0.0, 1.0{]} & Brake \\
2 & {[}0.5, 0.0, 0.0{]} & Turn Right \\
3 & {[}-0.5, 0.0, 0.0{]} & Turn Left \\
4 & {[}0.25, 0.5, 0.0{]} & Accelerate right \\
5 & {[}-0.25, 0.5, 0.0{]} & Accelerate Left \\
6 & {[}0.25, 0.0, 0.5{]} & Brake Right \\
7 & {[}-0.25, 0.0, 0.5{]} & Brake Left \\
8 & {[}0.0, 0.0, 0.0{]} & Coast \\
\hline
\end{tabular}
\caption{Mapping between a discrete action space and the continuous vehicle control commands represented using the normalized steering angle (steer:[-1,1]), the normalized throttle values (throttle:[0,1]) and the brake values (brake:[0, 1]) for training vehicle control policies}
\end{table}

\subsubsection{Reward Function}
\label{reward_func}

Each agent receives a reward given by $r_i(\mathbf{S^t}, E^t, a_i^t)$. Where the 
dependence on $E^t$, the environment state is used to signify that the reward function is 
also conditioned on the stochastic nature of the driving environment which includes 
weather, noisy communication channels etc. 

Similar to \cite{carla} we set the reward function to be a weighted sum 
of five terms: 1. distance traveled towards the goal $D$ in km, speed $V$ in km/h, 
collision damage $C$, intersection with sidewalk $SW \in [0, 1]$, and intersection 
with opposing lane $OL\in[0, 1]$

\begin{align}
    r_i = 1000\, (D_t-1 - D_t)+ 0.05\, (V_t - V_{t-1})- 0.00002\, \nonumber \\
    (C_t - C_{t-1}) -2\, (SW_t - SW_{t-1})-2\, (OL_t OL_{t-1}) \nonumber \\ 
    + \alpha + \beta
\end{align}%

Where, optionally, $\alpha$ is used to encourage/discourage 
cooperation/competitiveness among the agents and $\beta$ is used to shape the 
rewards under stochastic changes in the world state $E^t$ .

\subsection{Agent algorithm}

The MACAD-Agents used for this experiment is based on the IMPALA \cite{impala} architecture implemented using RLLib \cite{rllib} with the following hyper-parameters: 
\begin{lstlisting}[language=Python]
{
  # Discount factor of the MDP
    "gamma":0.99,
  # Number of steps after which the rollout gets cut
    "horizon":None,
  # Whether to rollout "complete_episodes" or "truncate_episodes"
    "batch_mode":"truncate_episodes",
  # Use a background thread for sampling (slightly off-policy)
    "sample_async":False,
  # Which observation filter to apply to the observation
    "observation_filter":"NoFilter",
  # Whether to LZ4 compress observations
    "compress_observations":False,
    "num_gpus":args.num_gpus
  # Impala specific config
  # From Appendix G in https://arxiv.org/pdf/1802.01561.pdf
  # V-trace params.
    "vtrace":True,
    "vtrace_clip_rho_threshold":1.0,
    "vtrace_clip_pg_rho_threshold":1.0,
  # System params.
  # Should be divisible by num_envs_per_worker
    "sample_batch_size":args.sample_bs_per_worker,
    "train_batch_size":args.train_bs,
    "min_iter_time_s":10,
    "num_workers":args.num_workers,
  # Number of environments to evaluate vectorwise per worker.
    "num_envs_per_worker":args.envs_per_worker,
    "num_cpus_per_worker":1,
    "num_gpus_per_worker":1,
  # Learning params.
    "grad_clip":40.0,
    "clip_rewards":True,
  # either "adam" or "rmsprop"
    "opt_type":"adam",
    "lr":6e-4,
    "lr_schedule":[
    [
      0,
      0.0006
    ],
    [
      20000000,
      0.000000000001
    ],
  # Anneal linearly to 0 from start 2 end
  ],
  # rmsprop considered
    "decay":0.99,
    "momentum":0.0,
    "epsilon":0.1,
  # balancing the three losses
    "vf_loss_coeff":0.5,
  # Baseline loss scaling
    "entropy_coeff":-0.01,
}
\end{lstlisting}

The agents use a standard deep CNN with the following filter configuration:
[[32, [8, 8], 4], [64, [4, 4], 2], [64, [3, 3], 1]] followed by a fully-connected layer for their policy networks.
In the shared-weights configuration, the agents share the weights of a 128-dimensional fully-connected layer that precedes the final action-logits layer.

\begin{figure}
  \centering
  \includegraphics[width=\linewidth]{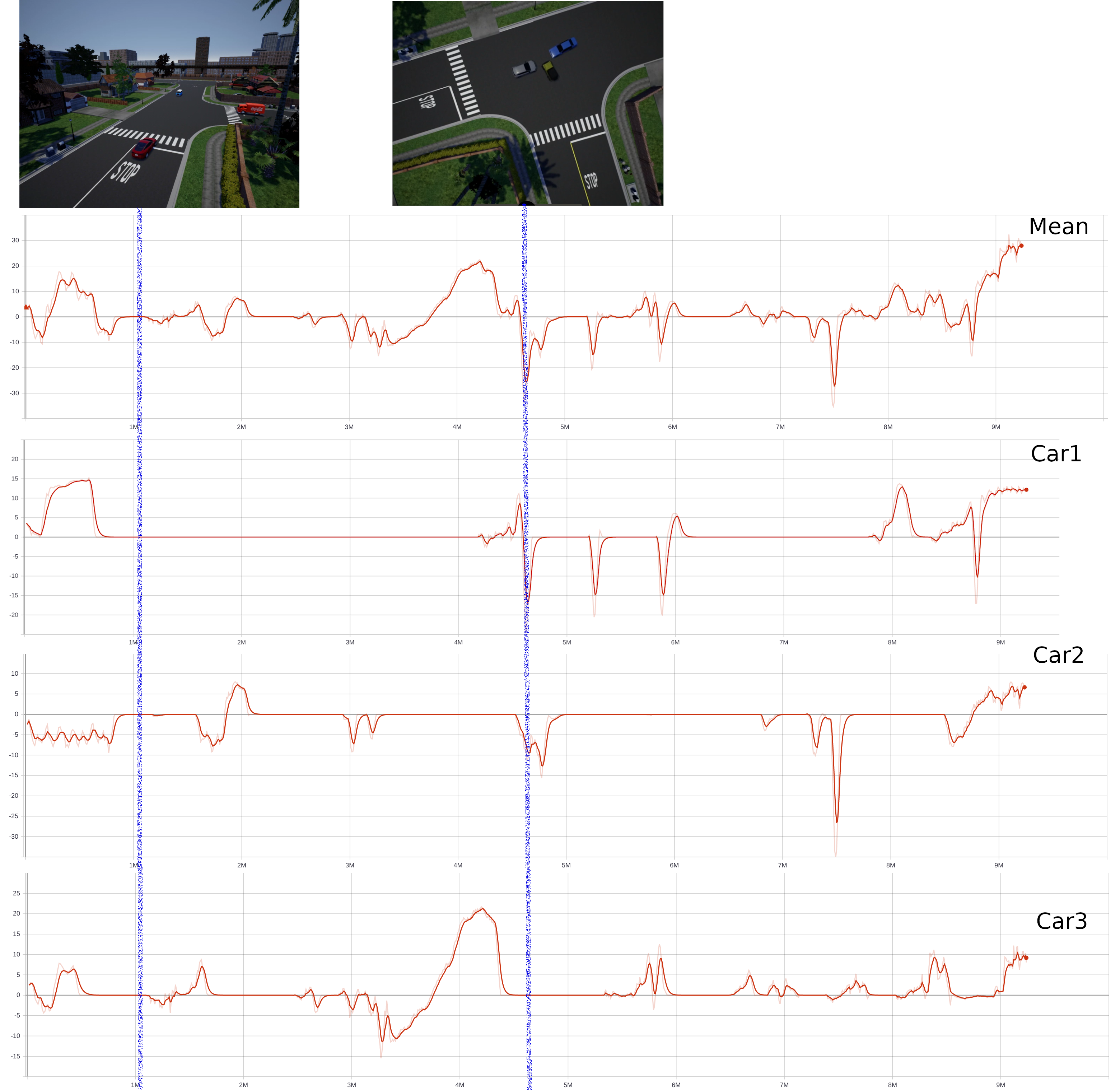}
  \caption{Figure shows the cumulative mean episodic rewards (legend:mean) and mean episodic rewards of car1 (legend:car1), car2 (legend:car2) and car3 (legend:car3). The blue vertical lines and the image at the top row indicate the states sampled during the corresponding training iteration (1.1M and 4.6M}.
  \label{fig:experiment-res}
\end{figure}

\subsection{Results}
The performance of the multi-agent system is shown in Figure \ref{fig:experiment-res}.

\end{document}